\documentclass[10pt,conference]{IEEEtran}
\usepackage{epsfig,rotating,setspace,latexsym,amsmath,epsf,amssymb,amsfonts,bm,theorem,cite,authblk, bbm,color, multirow, caption, subcaption, booktabs, cite}
\IEEEoverridecommandlockouts
\allowdisplaybreaks

\DeclareMathOperator*{\argmax}{arg\,max}

\title{How to Combat Reactive and Dynamic Jamming Attacks with Reinforcement Learning}

\author{Yalin E. Sagduyu}
\author{Tugba Erpek}
\author{Kemal Davaslioglu}
\author{Sastry Kompella}
\affil{\normalsize Nexcepta, Gaithersburg, MD, USA 
	\thanks{ This material is based upon work supported by the ASA(ALT) SBIR CCoE under Contract No. W51701-24-C-0151.}
} 

\date{}
\IEEEoverridecommandlockouts
\begin{document}
	
\maketitle
\vspace{-2cm}
\thispagestyle{empty}
	
\begin{abstract}
This paper studies the problem of mitigating reactive jamming, where a jammer adopts a dynamic policy of selecting channels and sensing thresholds to detect and jam ongoing transmissions. The transmitter–receiver pair learns to avoid jamming and optimize throughput over time (without prior knowledge of channel conditions or jamming strategies) by using reinforcement learning (RL) to adapt transmit power, modulation, and channel selection. Q-learning is employed for discrete jamming-event states, while Deep Q-Networks (DQN) are employed for continuous states based on received power. Through different reward functions and action sets, the results show that RL can adapt rapidly to spectrum dynamics and sustain high rates as channels and jamming policies change over time.
\end{abstract}
	
	
\section{Introduction}
The open wireless medium is inherently vulnerable to intentional interference, allowing malicious actors to degrade or even deny service across commercial and tactical networks. Conventional jammers broadcast noise persistently, overwhelming receivers and driving up errors, but reactive jammers pose an even greater threat by detecting legitimate transmissions and selectively injecting interference when needed. This adaptive behavior maximizes disruption while minimizing their power consumption and detectability. 
Thus, there is a critical need for agile, adaptive countermeasures that evolve alongside increasingly sophisticated jammers.

To preserve link reliability in both commercial and tactical settings, anti-jamming systems must dynamically infer jammer behavior and preemptively adapt their signaling strategies. Reinforcement learning (RL) is well suited to this task by framing the interaction as a sequential decision problem, where an RL agent can learn to select transmit parameters (e.g., power levels, modulation types, channels) that maximize long‐term throughput under adversarial interference. With exploration and continual policy refinement, RL methods naturally accommodate nonstationary jamming patterns, internalize the effect of each action on future states, and optimize trade-offs between immediate performance and future resilience. 
 
RL has been studied in various anti-jamming settings  \cite{abuzainab2019qos, wang2021jamming, 9771339, 9989422, 10287624,10773861} including reactive jammer scenarios\cite{8911228, 9751039,  10437052, 10812995, 10416893}.
Most RL studies against jammers treat each time slot as an independent decision epoch instead of a true Markov decision process (MDP) where actions shape future states. By using jammer models that ignore transmitter actions (treating the jammer as a stationary or simple Markov chain exogenous to the defender), the problem collapses to a bandit or block-MDP. The transmitter’s choice (e.g.,  channel or power) does not influence the jammer’s next state, which follows fixed transitions, so Q-learning and policy-gradient updates learn only one-step rewards under a static environment. This simplification can be misleading: in a genuine MDP, an optimal policy trades off immediate reward against long-term, action-dependent dynamics, whereas the common ``stateless" or exogenous-jammer formulation reduces the value function to a sum of independent one-step returns. Policies trained this way fail when a sophisticated jammer adapts to the transmitter.

This paper investigates mitigating reactive jamming in single and multi-channel environments by modeling the interaction with the jammer as an MDP, where the state evolves depending on transmit actions. The jammer intelligently updates channels and sensing thresholds to detect and disrupt transmissions, while the transmitter-receiver pair uses RL to adapt its power, modulation, and channel selection to minimize jamming impact and maximize throughput. Q-learning and Deep Q-Networks (DQN) are applied for discrete and continuous state spaces, respectively. The results show that RL enables rapid adaptation to changing jamming strategies and spectrum conditions, and sustain high throughput over time.
	
The remainder of the paper is organized as follows. Sec.~\ref{sec:system_model} describes the system model and optimization problem. Sec.~\ref{sec:pc} presents the RL algorithm and its performance under power control (PC). Joint power control and adaptive modulation (PCAM) is applied in Sec.~\ref{sec:pc_am}. The optimization problem is extended to multiple channels in Sec.~\ref{sec:mp}. The state is extended to a continuous case in Sec.~\ref{sec:css}. Sec.~\ref{sec:conclusion} concludes the paper.

\section{System Model} \label{sec:system_model}
	
We consider a wireless communication system operating in discrete slotted time where a transmitter communicates with its intended receiver in the presence of an adversarial jammer. 
The transmitter selects at each time slot \(t\) a transmit power and a modulation scheme from discrete sets. The transmit power \(P_{T}(t)\) at time $t$ is selected from a finite set \(\mathcal{P} = \{P_0, P_1, \dots, P_{K}\}\), where each \(P_k \in [0, P_{\mathrm{max}}]\). The modulation order \(M(t)\) at time $t$ is selected from a set \(\mathcal{M}\), representing M-ary QAM options. At time $t$, the wireless channel gain from the transmitter to the receiver is \(h_{T\!R}(t)\), the gain from the transmitter to the jammer is \(h_{T\!J}(t)\), and the gain from the jammer to the receiver is \(h_{J\!R}(t)\). The receiver experiences additive white Gaussian noise with variance \(\sigma_R^2\).
			
\noindent \textbf{Jammer.} The jammer uses an \emph{energy detector} (subject to noisy measurements) to decide whether to jam (with interference power \(P_I(t)\)) or not. The jammer jams with probability $P_D$, which is the probability of  detecting a transmission. Let $\chi^2_{2N}(\lambda)$ denote the noncentral chi–square distribution 
with $2N$ degrees of freedom and noncentrality $\lambda$. Then $P_D$ is given by
\begin{equation}
P_D(\tau(t)) = \Pr\!\left( \chi^2_{2N}(\lambda) > \tfrac{\tau(t)}{\sigma_J^2}\right),
\end{equation}
where $\tau(t)$ is the sensing threshold of energy detector, $N$ is the number of samples, 
$\lambda = N P_T(t) h_{T\!J}(t)/\sigma_J^2$, and $\sigma_J^2$ is the noise power at the energy detector.  \(\tau(t)\) switches between a low value \(\tau_{\mathrm{low}}\) and a high value \(\tau_{\mathrm{high}}\) depending on the previous jamming outcome. 
 The indicator $J_t = 1$ if the jammer transmits at time $t$ and 
 $0$, otherwise. The jammer’s energy‐detection threshold, \(\tau(t)\),  at time \(t\)  is chosen based on whether it jammed in the previous time slot:
\begin{equation}
\tau(t) =
\begin{cases}
	\tau_{\mathrm{high}}, & \text{if }J_{t-1} = 1,\\
	\tau_{\mathrm{low}}, & \text{if }J_{t-1} = 0,
\end{cases}
\end{equation}
or equivalently $\tau(t) = J_{t-1}\,\tau_{\mathrm{high}} \;+\;(1 - J_{t-1})\,\tau_{\mathrm{low}}$.
If the jammer cannot detect a signal, it reduces sensing threshold to increase the chance of detecting a signal in the next slot. Else, it increases sensing threshold to limit energy consumption in the next slot.
The jammer’s decision at time \(t\) is $J_{t} =	1$ with probability  $P_D(\tau(t))$, and $0$, otherwise.
When $\sigma_J^2 = 0$, the jamming decision is simplified as $J_{t} = 1$   if $P_T(t) h_{T\!J}(t) > \tau(t)$, $J_{t} =0$, if $P_T(t) h_{T\!J}(t) < \tau(t)$. Otherwise,  $J_{t} = 0$ or $1$ with probability $0.5$. Unlike exogenous jammer models, the jammer's future behavior depends on the transmitter’s actions.

\noindent \textbf{Transmitter.} 
The transmitter does not have direct access to the jammer’s behavior or sensing threshold. Instead, it observes feedback from the receiver, either a binary indication of whether the previous transmission was jammed or, more generally, the total received power. The transmitter’s decision-making is modeled as an MDP, where it selects actions based on the observed state and receives a corresponding reward.

\noindent \textbf{Reinforcement Learning:} The transmitter does not have access to the jammer’s internal model, hence cannot directly compute transition probabilities. Therefore, a model-free RL approach is adopted. The transmitter maintains a Q-value table \(Q(s, a)\) defined over states \(s \in \mathcal{S}\) and  actions \(a \in \mathcal{A}\). 

\noindent \textit{\(\epsilon\)-greedy Action Selection.} At time \(t\), given the current state \(s_t\), the agent selects action \(a_t\) according to an \(\epsilon\)-greedy policy:
\begin{equation}
a_t =
\begin{cases}
	\displaystyle \text{Uniform}(\mathcal{A}), & \text{with probability }\epsilon_t,\\[8pt]
	\displaystyle \argmax_{a\in A} Q_t(s_t,a), & \text{with probability }1 - \epsilon_t.
\end{cases}
\end{equation}
Here, \(\mathcal{A}\) is the discrete action set and \(Q_t(s,a)\) is the current estimate of the action–value function. At each $t$, the agent either explores or exploits. With probability $\epsilon_t$, it picks a uniformly random action; otherwise it picks the action with highest current Q-value for the current state. The agent gradually shifts from exploration toward exploitation over time, namely the exploration rate \(\epsilon_t\) decays after each episode according to $\epsilon_{t+1} = \max\bigl(\epsilon_{\mathrm{min}},\,\epsilon_t \cdot \epsilon_{\mathrm{decay}}\bigr)$, where \(\epsilon_{\mathrm{min}}\) is the minimum exploration probability, and \(\epsilon_{\mathrm{decay}}\in(0,1)\) is the decay rate. 

\noindent \textit{Q-learning Update Rule.} After taking action \(a_t\) in state \(s_t\), receiving reward \(r_{t+1}\) and transitioning to state \(s_{t+1}\), the Q-table is updated by one-step temporal-difference (TD) rule:
\begin{eqnarray}
&& \hspace{-1.4cm}Q_{t+1}(s_t,a_t) \nonumber
= Q_t(s_t,a_t)
\\&& + \alpha
\Bigl[
r_{t+1}
+ \gamma \,\max_{a'} Q_t(s_{t+1},a')
- Q_t(s_t,a_t)
\Bigr],
\end{eqnarray}
where \(\alpha\in(0,1]\) is learning rate, \(\gamma\in[0,1]\) is discount factor, and \(\max_{a'} Q_t(s_{t+1},a')\) estimates the optimal future value. 

The state is set as the jamming indicator in Secs.~\ref{sec:pc}, \ref{sec:pc_am}, and \ref{sec:mp}, and as the total received power in Sec.~\ref{sec:css}. For the single channel case, we  consider PC as action in Sec.~\ref{sec:pc} and PCAM as action in Sec.~\ref{sec:pc_am}. We extend the formulation to multiple channels by adding channel selection to the action space in Sec.~\ref{sec:mp}. We extend the discrete state to the continuous state  based on the total received power in Sec.~\ref{sec:css}.  

\noindent \textbf{Setting for Performance Evaluation:}  
While our approach is topology and channel-agnostic, we consider the following setting for performance evaluation. Transmitter (T) and receiver are at locations \((0,0)\) and \((1,0)\), respectively and jammer moves between position 1, closer to transmitter:  \((0,1)\) and position 2, closer to receiver: \((1,1)\). The channel gain between two nodes is modeled as path loss with path loss exponent 2 (free-space assumption). For both jammer positions, \(h_{T\!R} = 1\). For jammer position 1, \(h_{T\!J} = 1\) and \(h_{J\!R} = 0.5\). For jammer position 2, \(h_{T\!J} = 0.5\) and \(h_{J\!R} = 1\). The power of the jamming signal at the receiver is \(h_{J\!R}(t) P_I(t)\) when the jammer is active and \(P_I(t)\) is interference power. The receiver also experiences additive white Gaussian noise with variance \(\sigma_R^2\). 

The learning rate $\alpha$ determines how quickly the Q-values adapt to new observations, while the discount factor $\gamma$ controls the weight given to future rewards. The agent explores the action space using an $\epsilon$-greedy policy, where the exploration probability decays exponentially from $\epsilon_{\text{start}}$ to $\epsilon_{\text{final}}$ at rate $\epsilon_{\text{decay}}$. The training proceeds over $E = 20{,}000$ episodes, each consisting of $H = 200$ time steps. The Q-table $Q(s,a)$ is a $2 \times ((K+1) \times M)$ matrix (for two states and $(K+1) \times M$ actions), initialized to zero and updated online based on observed rewards and transitions. As the baseline, we assume that the transmitter is not adaptive and selects fixed power and the corresponding best modulation type as its action for a single channel. For multiple channels, the transmit parameters are fixed but the channel periodically changes in the baseline.

\section{Power Control} \label{sec:pc}

\noindent \textbf{State:} The state \(s_t \in \{0, 1\}\) indicates if the previous transmission was jammed (\(s_t = 1\)) or not (\(s_t = 0\)), namely $s_t = J_t$.

\noindent \textbf{Action.}
The transmitter's action is to select a discrete transmit power level, $P_T(t)$ from a uniformly spaced set $\mathcal{P}=[0, P_{\mathrm{max}}] $ (with resolution $1/K$ such that there are $K+1$ discrete power levels). Then, the action space is $\mathcal{A} = \{(P_k) : P_k \in \mathcal{P}\}$.
At time $t$, the transmitter chooses $a_t = P_T(t)$.

\noindent \textbf{Reward.} 
The reward is the Shannon rate, i.e., link capacity (bits/s/Hz)  $r_t = \log_2\!\bigl(1 + \mathrm{SINR}_t\bigr)$, where the signal-to-interference-plus-noise ratio (SINR) at the receiver is
\begin{equation}
	\mathrm{SINR}_t =
	\begin{cases}
		\dfrac{h_{T\!R}(t)P_T(t)}{h_{J\!R}(t) P_I(t) + \sigma_R^2}, & \text{if } J_t = 1,\\[1em]
		\dfrac{h_{T\!R}(t) P_T(t)}{\sigma_R^2}, & \text{if } J_t = 0,
	\end{cases}
\end{equation}
where $t$ denotes the time instance, $P_I(t) = P_I$ and $h_{T\!R}(t)=1$ for numerical results, and $h_{T\!R}(t)$, $h_{J\!R}(t)$, and $P_T(t)$ may change over time. The transmitter learns to select the power for each observed state to maximize the long-term reward. 

\begin{table}[ht]
	\footnotesize
	\centering
	\caption{Parameters for PC in single channel case.}
	\label{tab:jam_env_params}
	\begin{tabular}{ll}
		\toprule
		\textbf{Parameter (Symbol)} & \textbf{Value} \\
		\midrule
		Maximum transmit power ($P_{\mathrm{max}}$) & 1.0 \\
		Low jamming threshold ($\tau_{\mathrm{low}}$) & 0.2 \\
		High jamming threshold ($\tau_{\mathrm{high}}$) & 0.4 \\
		Number of power levels ($K+1$) & 101 \\
        Number of samples ($N$) & 1 \\
		Episode length & 200 \\
		Total episodes & 20000 \\
		Interference power ($P_{I}$) & 100 \\
		Receiver noise variance ($\sigma_R^2$) & 0.1 \\
		Jammer–transmitter gain ($h_{T\!J}$) & [0.5,\,1.0,\,0.5,\,1.0] \\
		$h_{T\!J}$ changes (episodes) & [0,5k),\,[5k,10k),\,[10k,15k),\,[15k,20k) \\
		Learning rate ($\alpha$) & 0.1 \\
		Discount factor ($\gamma$) & 0.95 \\
		Initial exploration ($\epsilon_{\mathrm{start}}$) & 1.00 \\
		Final exploration ($\epsilon_{\mathrm{final}}$) & 0.01 \\
		Exploration decay ($\epsilon_{\mathrm{decay}}$) & 0.999 \\
		\bottomrule
	\end{tabular}
\end{table}

\begin{figure}[ht!]
		\centering
		\includegraphics[width=0.64\columnwidth]{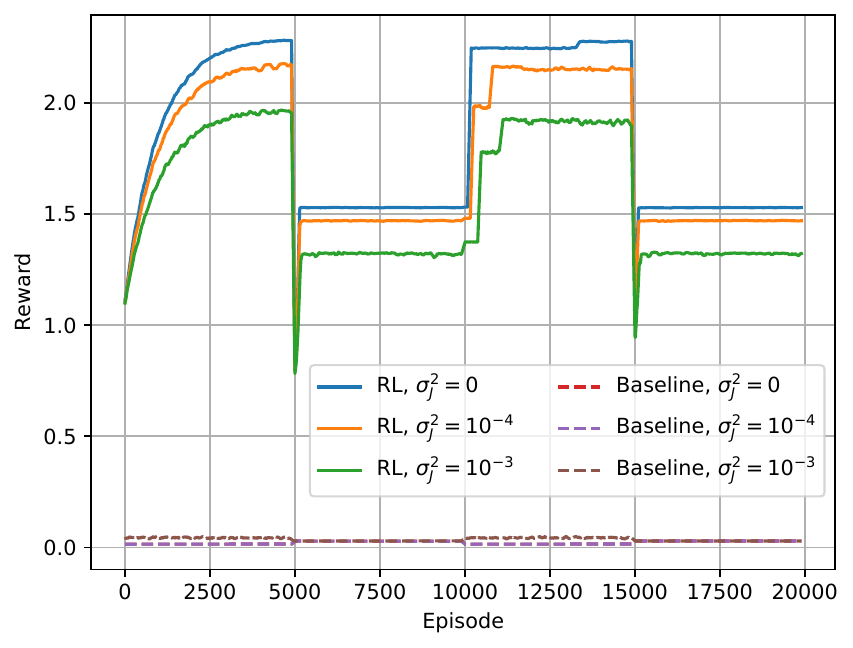}
		\vspace{-0.05cm}
		\caption{Reward over episodes for PC in single channel case.}
		\label{fig:SC_Q_pc_reward}
\end{figure}

\begin{figure}[ht!]
	\vspace{-0.15cm}
	\centering
	\includegraphics[width=0.64\columnwidth]{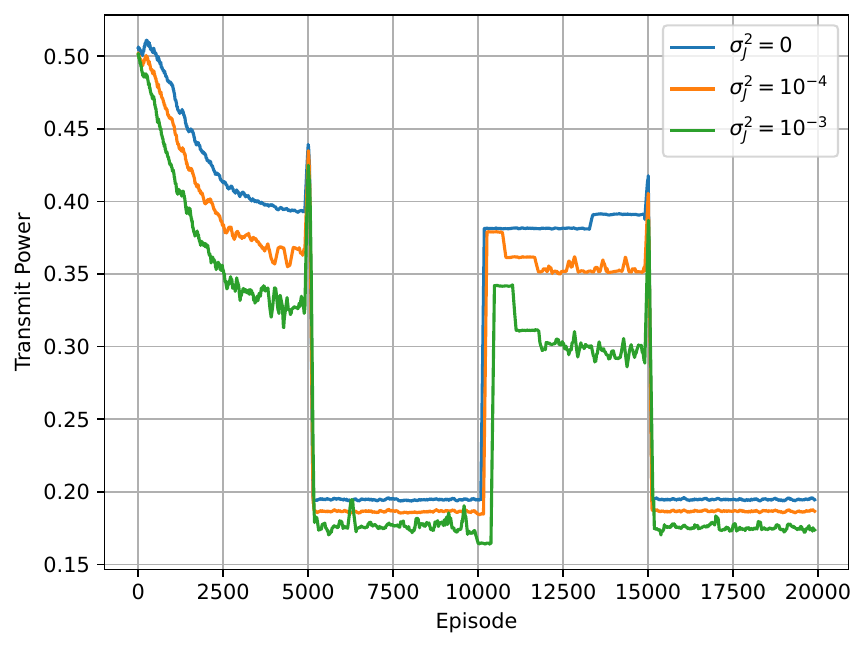}
	\vspace{-0.05cm}
	\caption{Transmit power over episodes for PC in single channel case.}
	\label{fig:SC_Q_pc_power}
\end{figure}

\begin{figure}[ht!]
	\vspace{-0.15cm}
	\centering
	\includegraphics[width=0.64\columnwidth]{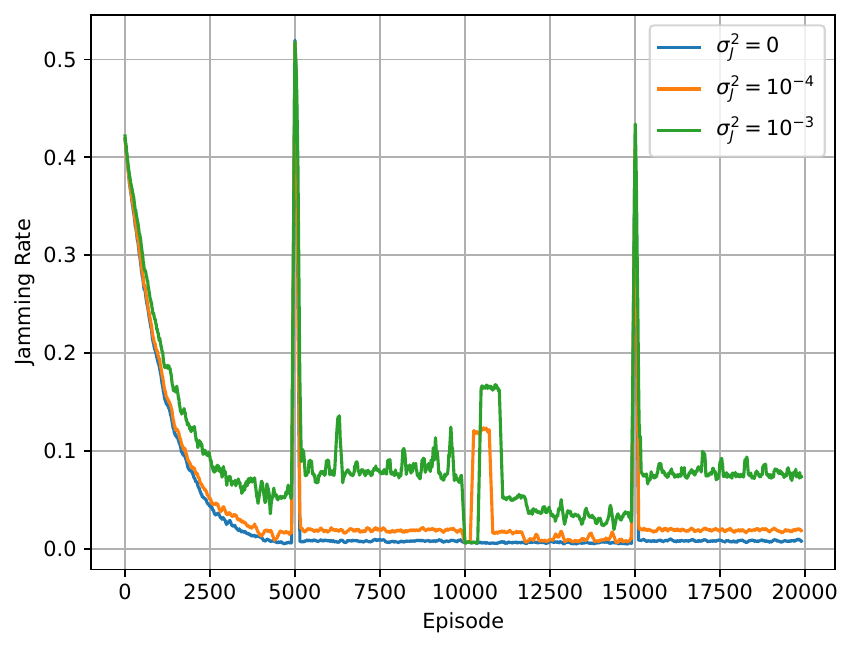}
	\vspace{-0.05cm}
	\caption{Jamming rate over episodes for PC in single channel case.}
	\label{fig:SC_Q_pc_jamrate}
    \vspace{-0.25cm}
\end{figure}

 In numerical results, parameters follow from Table~\ref{tab:jam_env_params}. Transmit power \(P_T\) is selected as $P_k = \frac{k}{K}$ for $0 \leq k \leq K$, where $K = 100$. Fig.~\ref{fig:SC_Q_pc_reward} shows the reward over episodes and Table \ref{tab:SC_Q_pc_reward} shows the aggregated reward. The reward is higher  when $\sigma_J^2$ is lower. For the initial (low) $h_{T\!J}$, RL learns how to maximize its reward over time by adjusting \(P_T\) (as shown in Fig.~\ref{fig:SC_Q_pc_power}) and consequently reducing the jamming rate, i.e., fraction of time slots in which the transmission is jammed (as shown in Fig.~\ref{fig:SC_Q_pc_jamrate}). When $h_{T\!J}$ increases, the jamming rate spikes and the reward drops suddenly. Then RL quickly adapts to the change by decreasing \(P_T\) and learning to avoid the jammer, and the reward increases back quickly. As $h_{T\!J}$ changes over time, there are short-term peaks for jamming rate, which is quickly mitigated in a small number of episodes. The transmitter effectively learns to optimize \(P_T\) while maximizing the reward such that the receiver signal power at the jammer remains below the sensing threshold (unknown to the transmitter). The baseline using the fixed maximum power is not effective against the jammer, yielding smaller rates.

\begin{table}[h!]
	\footnotesize
	\centering
    \vspace{-0.15cm}
	\caption{Total reward for PC in a single channel under different   $\sigma_J^2$.}
	\label{tab:SC_Q_pc_reward}
	\vspace{0.25cm}
	\begin{tabular}{c|c|c|c|c|c}
		\hline
		\textbf{Policy} & $\sigma_J^2$ & \textbf{Total reward} & \textbf{Policy} & $\sigma_J^2$ & \textbf{Total reward} \\
		\hline
		 & $0$       & 36552.44 & & $0$       & 428.54\\ 
		RL  & $10^{-4}$     & 34849.52 & Fixed  & $10^{-4}$     & 428.54 \\  
		 & $10^{-3}$     & 31264.54 &  &$10^{-3}$     & 717.91 \\ 
		\bottomrule

	\end{tabular}
\end{table}

\section{Joint Power Control and Adaptive Modulation} \label{sec:pc_am}
We add the capability of selecting modulation scheme in addition to transmit power for the transmitter. State is the same as in Sec.~\ref{sec:pc}. Action and reward are updated as follows.

\noindent \textbf{Action.}  The transmitter's action is to select not only a discrete transmit power, but also one out of $M$ modulation schemes, $M(t)$, from $\mathcal{M}$. Formally, the action space is defined as
\begin{equation}
\mathcal{A} = \{(P_k, M_l) : P_k \in \mathcal{P},\; M_l \in \mathcal{M}\}.
\end{equation}
At time \(t\), the transmitter chooses \(a_t = (P_T(t), M(t))\).

\noindent \textbf{Reward.}			
Based on the SINR, the reward is the achievable (uncoded) throughput under the chosen modulation, given by 
\begin{equation} \label{eq:ber}
r_t = \log_2(M(t)) \times \left(1 - \mathrm{BER}(M(t); \mathrm{SINR}_t)\right), 
\end{equation}
where \(\mathrm{BER}(M; \mathrm{SINR}_t)\) is the bit error rate of the modulation \(M\)-QAM under SINR \(\mathrm{SINR}_t\). This expression captures both the spectral efficiency and the reliability of communication.
			
\begin{figure}[ht!]
	\vspace{-0.25cm}
	\centering
	\includegraphics[width=0.64\columnwidth]{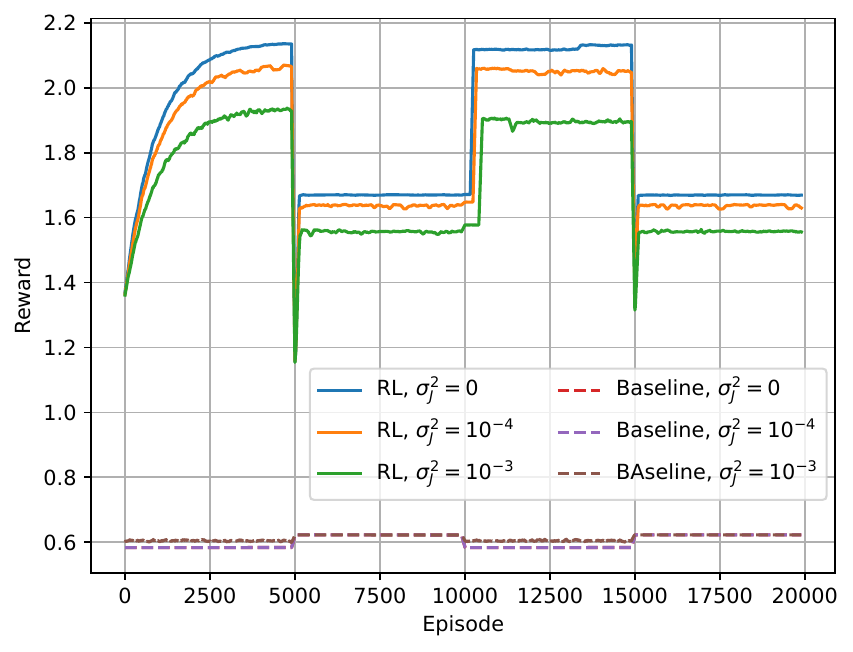}
	\vspace{-0.05cm}
	\caption{Reward over episodes for PCAM in single channel case.}
	\label{fig:SC_Q_pcam_reward}
\end{figure}

Over time, the transmitter learns to select the best power and modulation pair for each observed state to maximize long-term reward as jamming and channel changes. The transmit power \(P_T\) is selected as \(P_k = \frac{k}{K} \) for $0 \leq k \leq K$, where $K = 100$. The modulation order \(M\) is selected from a set \(\mathcal{M} = \{2, 4, 8, 16, 32, 64\}\), representing M-ary QAM options. Fig.~\ref{fig:SC_Q_pcam_reward} shows reward over episodes and  Table \ref{tab:SC_Q_pcam_reward} shows the aggregated reward. The reward  increases as $\sigma_J^2$ decreases. 

\begin{table}[ht!]
	\vspace{-0.4cm}
	\footnotesize
	\centering
	\caption{Total reward for PCAM in a single channel under different  $\sigma_J^2$.}
	\label{tab:SC_Q_pcam_reward}
	\vspace{0.25cm}
	\begin{tabular}{c|c|c|c|c|c}
		\hline
		\textbf{Policy} & $\sigma_J^2$ & \textbf{Total reward} & \textbf{Policy} & $\sigma_J^2$ & \textbf{Total reward} \\
		\hline
		& $0$       & 37053.69 & & $0$       & 12059.80 \\ 
		RL  & $10^{-4}$     & 36055.70 & Fixed  & $10^{-4}$     & 12059.80\\  
		& $10^{-3}$     & 33905.70 & & $10^{-3}$     & 12272.49 \\ 
		\hline	 
	\end{tabular}
\end{table}

\begin{figure}[t!]
	\centering
	\includegraphics[width=0.64\columnwidth]{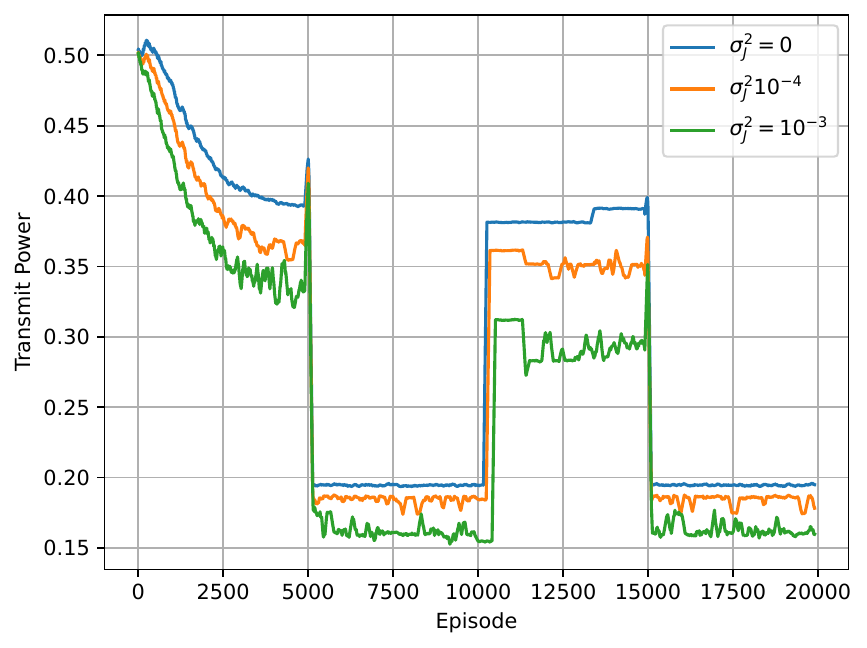}
	\vspace{-0.05cm}
	\caption{Transmit power over episodes for PCAM in single channel case.}
	\label{fig:SC_Q_pcam_power}
    	\vspace{-0.15cm}
\end{figure}

\begin{figure}[ht!]
	\centering
	\includegraphics[width=0.64\columnwidth]{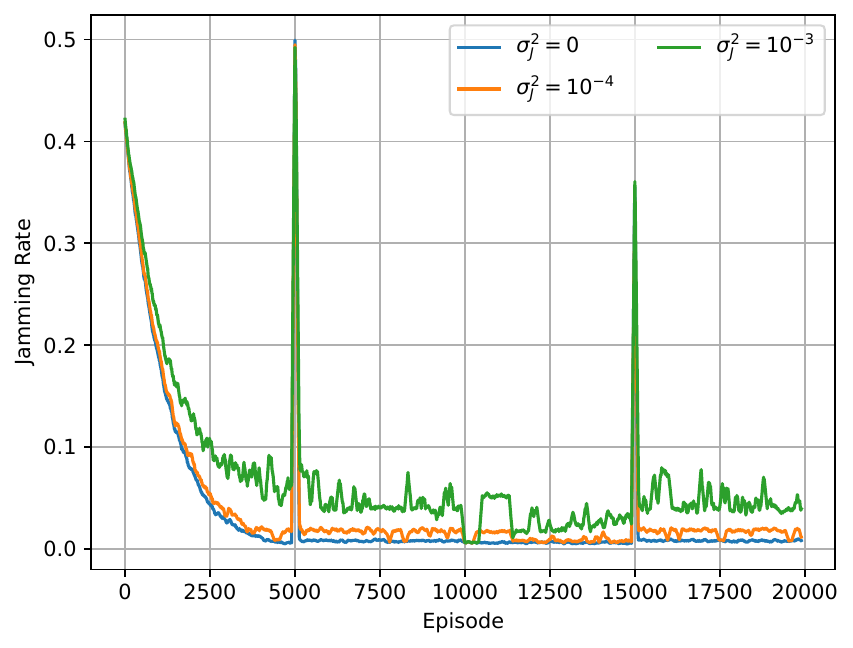}
	\vspace{-0.05cm}
	\caption{Jamming rate over episodes for PCAM in single channel case.}
    \vspace{-0.15cm}
	\label{fig:SC_Q_pcam_jamrate}
\end{figure}

\begin{figure}[h!]
	\centering
	\includegraphics[width=0.64\columnwidth]{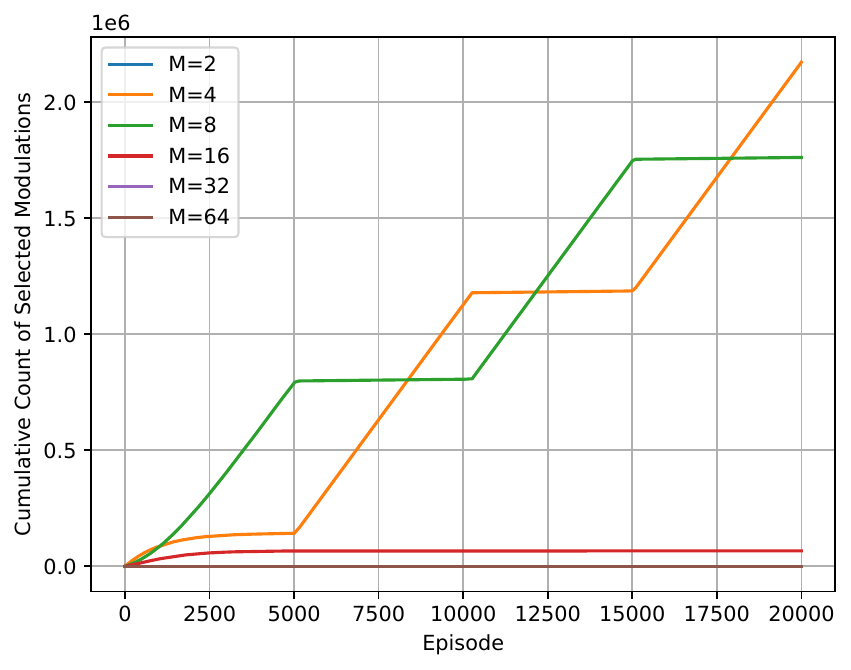}
	\vspace{-0.05cm}
	\caption{Count of modulation types for PCAM in single channel case.}
    \vspace{-0.20cm}
	\label{fig:SC_Q_pcam_mod}
\end{figure}

Under PCAM, the reward increases compared to the PC case. Figs.~\ref{fig:SC_Q_pcam_power}, \ref{fig:SC_Q_pcam_jamrate}, and \ref{fig:SC_Q_pcam_mod} show transmit power, jamming rate, and count of modulation types over episodes, respectively. A higher modulation ($M=8$) is mostly selected when $h_{T\!J}$ is low and the transmit power is increased, whereas a lower modulation ($M=4$) is mostly selected when $h_{T\!J}$ is high and the transmit power is reduced to optimize the effective throughput for the resulting SINR. The baseline using fixed maximum power and corresponding modulation is ineffective against the jammer (the reward is higher compared to the baseline PC, since the effective throughput does not fully diminish with low SINR compared to the Shannon rate).

\section{Extension to Multiple Channels} \label{sec:mp}

We consider multiple ($N$) channels from set $\mathcal{C}$. At any given time $t$, the transmitter selects channel (frequency) $c_T(t)$ and transmits on this channel. On the other hand, the jammer selects channel $c_J(t)$, senses that channel, and jams it if it detects any transmission. We consider the same state formulation as in previous section and update action and reward as follows.

\noindent \textbf{State:} The state $s_t$ $\in \mathcal{S}$ $=$ $\{0, 1\}$ captures whether the previous transmission was jammed (\(s_t = 1\)) or not (\(s_t = 0\)), namely $s_t = d_t$, where $d_t = \mathbf{1}\bigl(c_T(t) = c_J(t), J_t = 1\bigr)$ and $\mathbf{1}(\cdot)$ is the indicator function.

\noindent \textbf{Action.} We consider PCAM and add channel selection to the action space. This way, the action space is extended to
\begin{equation}
	\mathcal{A} = \{(c_j, P_k, M_l) : c_j \in \mathcal{C}, \; P_k \in \mathcal{P},\; M_l \in \mathcal{M}\}.
\end{equation}
At time \(t\), the transmitter chooses \(a_t = (c_T(t), P_T(t), M(t))\).

\noindent \textbf{Reward.} The  SINR at the receiver is given by
\begin{equation}
	\mathrm{SINR}_t =
	\begin{cases}
		\dfrac{h_{T\!R}(t) P_T(t)}{h_{J\!R}(t) P_I(t) + \sigma_R^2}, & \text{if } c_T(t) = c_J(t),\\[1em]
		\dfrac{h_{T\!R}(t)P_T(t)}{\sigma_R^2}, & \text{if } c_T(t) \neq c_J(t).
	\end{cases} \label{eq:SINR_mc}
\end{equation}

The reward is given by (\ref{eq:ber}) for the SINR in (\ref{eq:SINR_mc}).

\noindent \textbf{Jammer.} The jammer updates its channel and threshold as
\begin{eqnarray}
	&&
	c_J(t+1) =
	\begin{cases}
		c_J(t),               &\text{w.p. }p^{k},\\
		c \in \mathcal{C} \setminus c_J(t),           &\text{w.p. } \frac{1-p^k}{N-1},
	\end{cases}
	\\
	&&
	\tau_{t+1} =
	\begin{cases}
		\tau^k,  &\text{if } c_J(t+1) = c_J(t),\\
		\bar{\tau}^k  & \text{if }c_J(t+1) \neq c_J(t),
	\end{cases}
\end{eqnarray}
when $d_t = k$, where $p^{k} = p, \tau^k =\tau_{\mathrm{high}}, \bar{\tau}^k = \tau_{\mathrm{low}}$ if $k=1$, and $p^{k} = q, \tau^k =\tau_{\mathrm{low}}, \bar{\tau}^k = \tau_{\mathrm{high}}$ if $k=0$. We assume $p>0.5$ and $q<0.5$ such that the jammer more likely stays on the channel where it detected and jammed a transmission, and more likely moves to another channel otherwise. Mimicking the threshold update in the single channel case, the jammer becomes more aggressive and reduces the sensing threshold when it does not detect a transmission and decides to stay on the same channel. For a symmetric case, the jammer becomes aggressive and reduces its sensing threshold when it moves to a new channel after detecting and jamming a transmission.     

\begin{figure}[ht!]
	\centering
	\includegraphics[width=0.64\columnwidth]{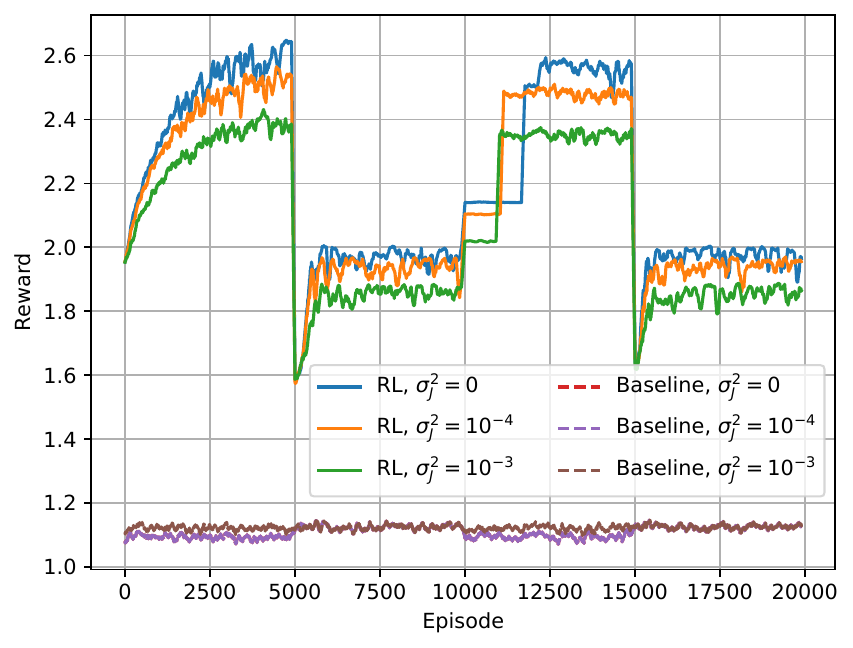}
	\vspace{-0.05cm}
	\caption{Reward over episodes in multi-channel case.}
	\label{fig:MC_Q_pcam_reward}
    	\vspace{-0.15cm}
\end{figure}

\begin{figure}[ht!]
	\vspace{-0.25cm}
	\centering
	\includegraphics[width=0.64\columnwidth]{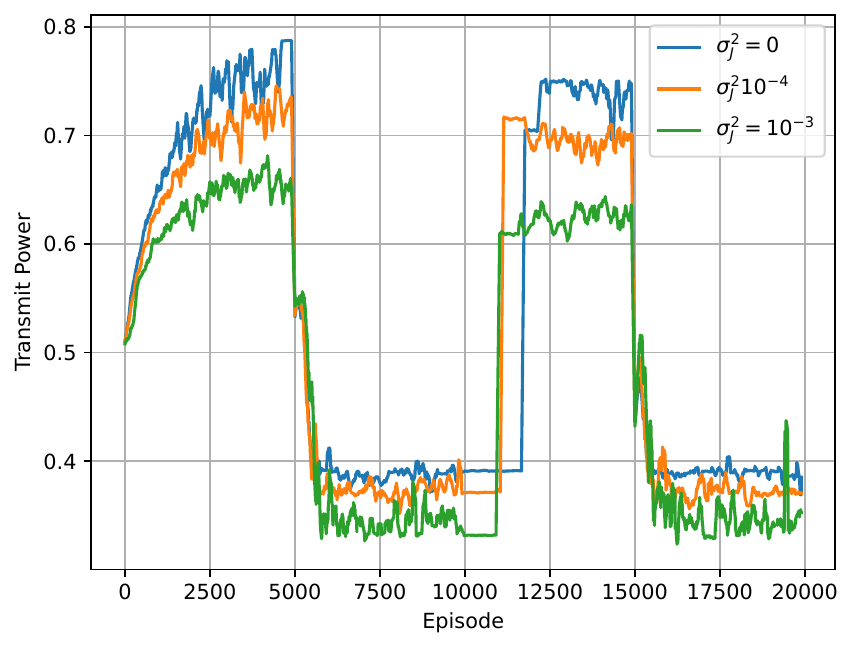}
	\vspace{-0.05cm}
	\caption{Transmit power over episodes in multi-channel case.}
	\label{fig:MC_Q_pcam_power}
    \vspace{-0.15cm}
\end{figure}

\begin{figure}[ht!]
\vspace{-0.25cm}
	\centering
	\includegraphics[width=0.64\columnwidth]{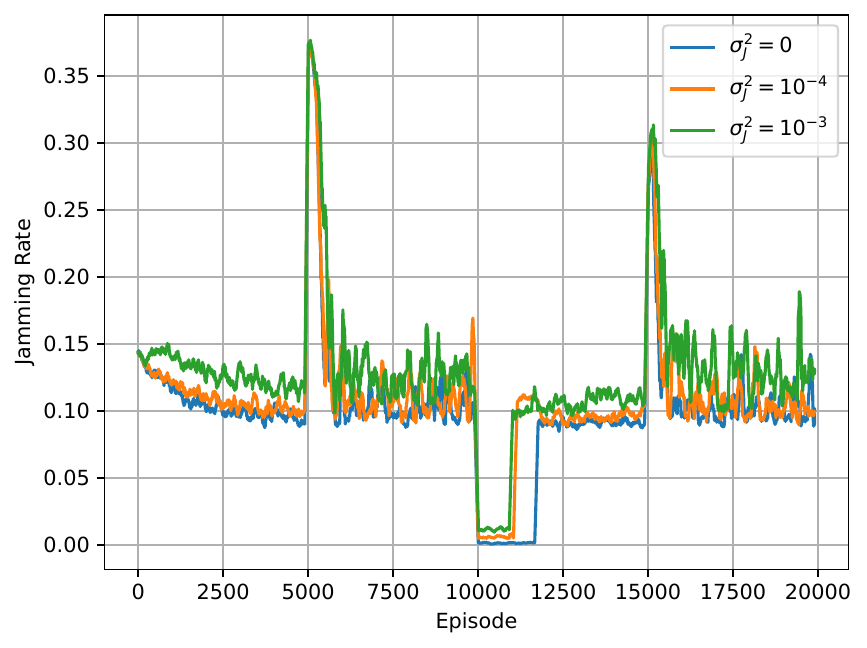}
	\vspace{-0.05cm}
	\caption{Jamming rate over episodes in multi-channel case.}
	\label{fig:MC_Q_pcam_jamrate}
    \vspace{-0.15cm}
\end{figure}

\begin{figure}[ht!]
	\centering
	\includegraphics[width=0.64\columnwidth]{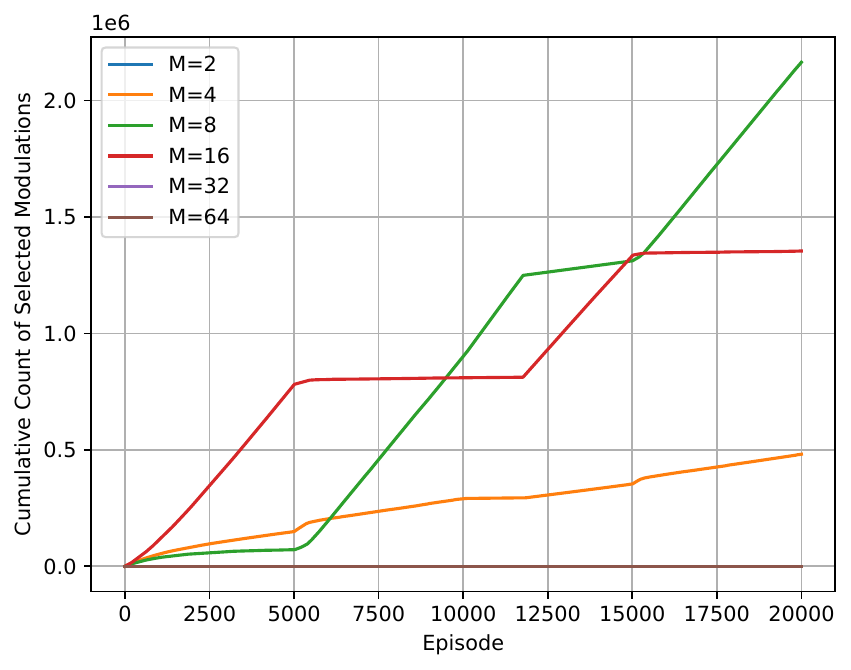}
	\vspace{-0.25cm}
	\caption{Count of modulation types in multi-channel case.}
    \vspace{-0.05cm}
	\label{fig:MC_Q_pcam_mod}
\end{figure}

For performance evaluation, we set $N=2$, $p=0.8$, and $q=0.2$. Figs.~\ref{fig:MC_Q_pcam_reward}, \ref{fig:MC_Q_pcam_power}, \ref{fig:MC_Q_pcam_jamrate}, and \ref{fig:MC_Q_pcam_mod} show the reward, transmit power, jamming rate, and count of modulation types over episodes, respectively, and Table~\ref{tab:C_Q_pcam_reward} shows the aggregated reward that is higher than the single channel case. While jamming can happen more due to randomness in channel decisions, it is likely to avoid the jammer by changing channels so when any idle channel is found, it is possible to increase transmit power more than the single channel and the reward improves accordingly. In the meantime, higher modulations (e.g., $M=16$) are selected to support higher rates.  

\begin{table}[h!]
	\footnotesize
	\centering
	\caption{Total reward in multi-channel case under different $\sigma_J^2$.}
	\label{tab:C_Q_pcam_reward}
	\vspace{0.25cm}
	\begin{tabular}{c|c|c|c|c|c}
		\hline
		\textbf{Policy} & $\sigma_J^2$ & \textbf{Total reward} & \textbf{Policy} & $\sigma_J^2$ & \textbf{Total reward} \\
		\hline
		& $0$       & 43736.50 & & $0$       & 22195.21\\  
		RL  & $10^{-4}$     & 43057.46 & Fixed  & $10^{-4}$     & 22195.21\\ 
		& $10^{-3}$     & 41183.54 & & $10^{-3}$     & 22461.99 \\
		\hline	 
	\end{tabular}
    \vspace{-0.1cm}
\end{table}

\section{Extension to Continuous State Space} \label{sec:css}
In the discrete‐state version of the MDP problem, the agent’s state was a binary indicator of whether the previous transmission was jammed.  In many practical scenarios, however, the receiver may not be able to detect jamming reliably at every time slot. Next, we replace this binary state by a continuous observation, namely the total received power. 

\noindent \textbf{Continuous State.} At time $t$, the state is  the received power: 
\begin{equation}
s_t =  h_{T\!R}(t) \,P_T(t) + J_t h_{J\!R}(t) P_I(t)  + \sigma_R^2.
\end{equation}
	
\noindent \textit{Function Approximation.} We use a neural network $Q(s,a;\theta)$ parameterized by weights $\theta$
to approximate the optimal Q-function over a continuous state \(s\) and discrete actions \(a \in \mathcal{A}\), namely \( Q(s,a;\,\theta)\approx Q^*(s,a)\). A separate target network
\( Q(s,a;\theta^{-}) \) keeps delayed weights \(\theta^{-}\).

\noindent \textit{\(\epsilon\)-greedy Policy.} At each time step $t$, with probability $\epsilon_t$ the agent chooses a random action, otherwise it selects the action that maximizes the current Q-network’s output. After each episode,  $\epsilon_t$ is decayed multiplicatively down to a minimum value, as also discussed in Q-learning.

\noindent \textit{Experience Replay.} Each transition \((s_t,a_t,r_{t+1},s_{t+1},e_t)\), where \(e_t\in\{0,1\}\) indicates the end of the episode, is stored in a replay buffer \(\mathcal{D}\).  At each learning step, we sample a minibatch \(\{(s_i,a_i,r_{i+1},s_{i+1},e_i)\}_{i=1}^B\) uniformly from \(\mathcal{D}\).

\noindent \textit{Temporal-Difference Target.} For each sample, the target is computed as $y_i = r_{i+1} + \gamma\,(1 - e_i)\,\max_{a'}Q\bigl(s_{i+1},a';\,\theta^{-}\bigr)$.

\noindent \textit{Loss and Gradient Descent.} The mean squared error between the target $\{y_i\}$ and the current estimate of the network, $\{Q(s_i,a_i;\theta)\}$, namely the mean-squared Bellman error 
\begin{equation}
L(\theta)
= \frac{1}{B}\sum_{i=1}^B
\Bigl[
y_i - Q(s_i,a_i;\theta)
\Bigr]^2,
\end{equation}
is minimized and network parameters are updated by stochastic gradient descent with learning rate $\eta$: $\theta \;\leftarrow\;\theta - \;\eta\,\nabla_{\theta}\,L(\theta)$.

\noindent \textit{Target Network Update.} Every \(C\) gradient steps, we copy  $\theta$  into  $\theta^{-}$ to stabilize training. 
	
\begin{figure}[ht!]
	\vspace{-0.05cm}
	\centering
	\includegraphics[width=0.64\columnwidth]{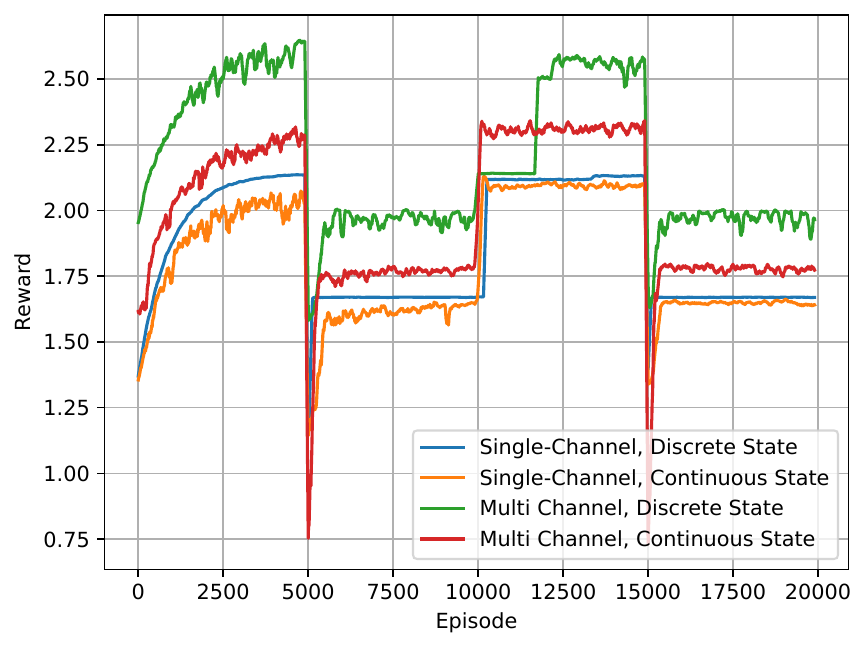}
	\vspace{-0.05cm}
	\caption{Reward over episodes under discrete and continuous states.}
	\label{fig:SC_MC_Cont_pcam_reward}
\end{figure}

\begin{figure}[ht!]
	\vspace{-0.4cm}
	\centering
	\includegraphics[width=0.64\columnwidth]{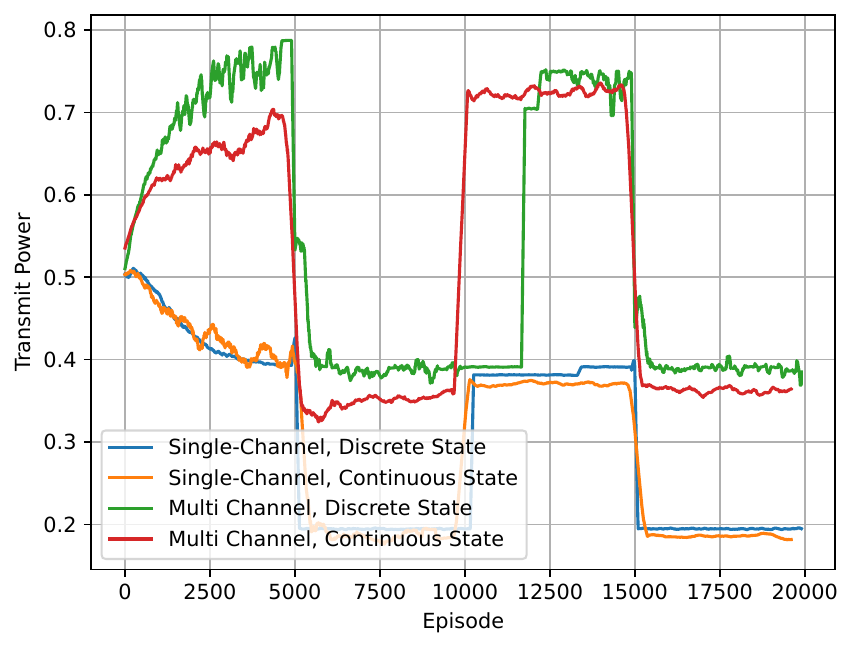}
	\vspace{-0.05cm}
	\caption{Transmit power over episodes under discrete and continuous states.}
	\label{fig:SC_MC_Cont_pcam_power}
\end{figure}

\begin{figure}[ht!]
	\vspace{-0.4cm}
	\centering
	\includegraphics[width=0.64\columnwidth]{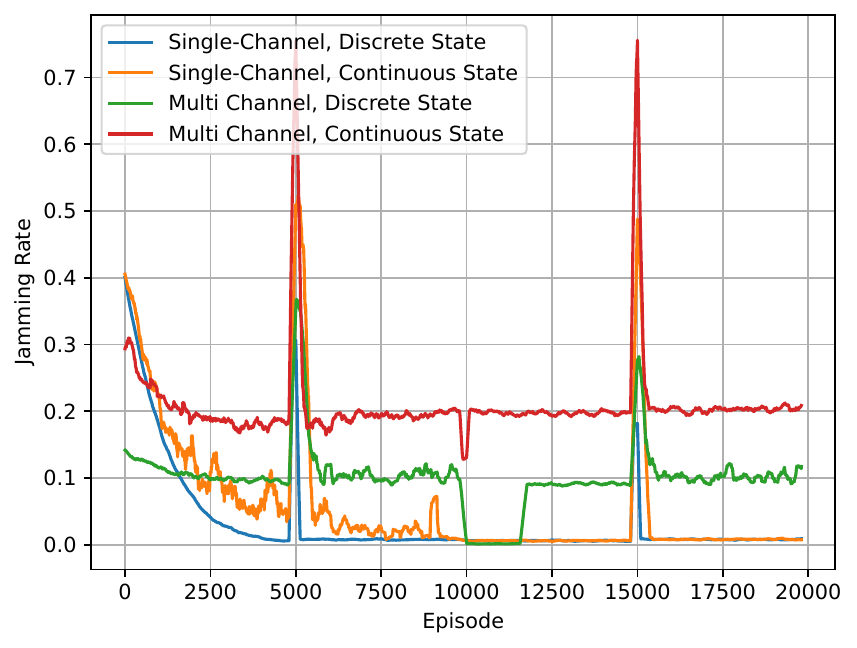}
	\vspace{-0.05cm}
	\caption{Jamming rate over episodes under discrete and continuous states.}
	\label{fig:SC_MC_Cont_pcam_jamrate}
\end{figure}

\begin{figure}[ht!]
	\centering
	\includegraphics[width=0.64\columnwidth]{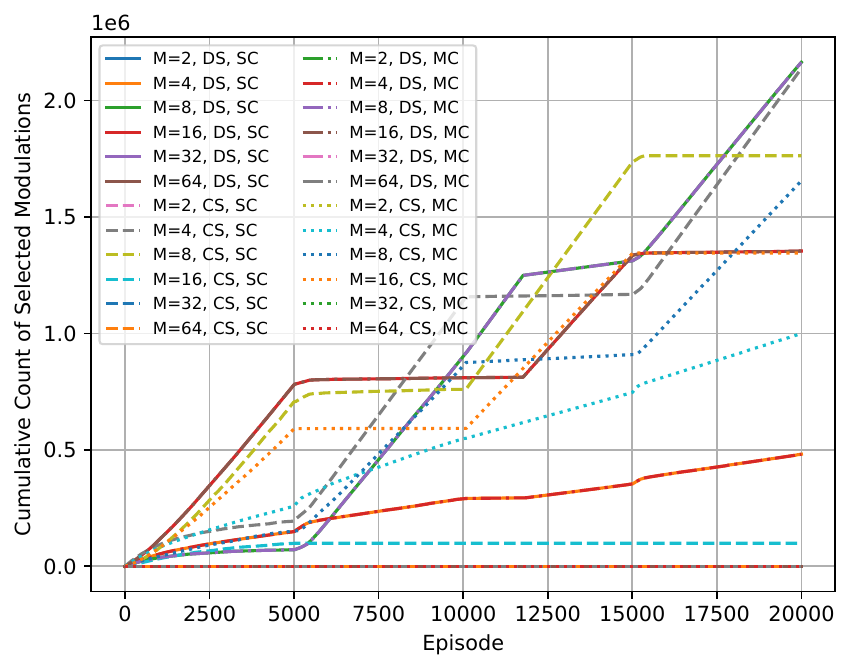}
	\vspace{-0.05cm}
	\caption{Count of modulation types under discrete state (DS) and continuous state (CS) in single channel (SC) and multi-channel (MC) cases.}
	\label{fig:SC_MC_Cont_pcam_mod}
\end{figure}

We evaluate the performance for PCAM $\sigma_J^2 = 0$. 
Hyperparameter tuning leads to the deep neural network setting with two hidden layers size of 64, each followed by ReLU activation function, replay buffer capacity of 100,000, and mini‐batch size of 64 in addition to learning parameters from Table~\ref{tab:jam_env_params}. Comparing single and multi-channel cases, Figs.~\ref{fig:SC_MC_Cont_pcam_reward},  \ref{fig:SC_MC_Cont_pcam_power}, \ref{fig:SC_MC_Cont_pcam_jamrate}, and \ref{fig:SC_MC_Cont_pcam_mod} show the reward, transmit power, jamming rate, and count of modulation type over episodes, respectively, and Table \ref{tab:SC_MC_Cont_pcam_reward} shows the aggregated reward. With the continuous state, the reward is lower (as DQN approximates the Q-table) and there are more fluctuations in transmit decisions and performance results (due to higher dynamics in received power than jamming indicator). Accordingly, transmit powers are lower to avoid the jammer, while jamming rates are higher, and lower modulations are selected to match the lower SINR.

\begin{table}[h!]
	\vspace{-0.15cm}
	\footnotesize
	\centering
	\caption{Total reward under discrete and continuous states.}
	\label{tab:SC_MC_Cont_pcam_reward}
	\begin{tabular}{c|c|c}
		\hline
		\textbf{State Type $\sigma_J^2$} & Channel Type  &\textbf{Total reward} \\
		\hline
		Discrete State       & Single Channel & 37053.69 \\
		Continuous State     & Single Channel & 35972.87 \\
		Discrete State       & Multi-Channel  & 43736.50 \\
		Continuous State     & Multi-Channel  & 39439.82 \\
		\hline
	\end{tabular}
		\vspace{-0.3cm}
\end{table}

\section{Conclusion} \label{sec:conclusion}
	In this paper, we studied the challenge of countering reactive and dynamic jamming in single and multi-channel settings, where a jammer dynamically selects channels and sensing thresholds to disrupt transmissions. In response, a transmitter-receiver pair uses RL to optimize throughput by adjusting transmission power, modulation, and channel selection without prior knowledge of channel conditions or jamming strategies. The study employs Q-learning and DQN for discrete and continuous state spaces, respectively, and demonstrates that RL enables effective adaptation to evolving jamming tactics and channel conditions, sustaining high throughput over time.

\vspace{-0.01cm}
\bibliographystyle{IEEEtran}

\bibliography{references_v3}

\end{document}